%% file: fate.tex
\documentclass[10pt,twocolumn,letterpaper]{article}

\usepackage{cvpr}
\usepackage{times}
\usepackage{epsfig}
\usepackage{graphicx}
\usepackage{amsmath}
\usepackage{amssymb}

\usepackage{multirow}
\usepackage[table,xcdraw]{xcolor}

\usepackage[pagebackref=true,breaklinks=true,letterpaper=true,colorlinks,bookmarks=false]{hyperref}

\cvprfinalcopy 

\def\cvprPaperID{****} 

\ifcvprfinal\fi
\begin{document}

\title{Implications of Computer Vision Driven Assistive Technologies Towards Individuals with Visual Impairment\thanks{We thank NSERC, Canada Research Chairs program, and
Microsoft.}}

\author{Linda Wang and Alexander Wong \\
Waterloo Artificial Intelligence Institute \\
University of Waterloo\\
{\tt\small \{ly8wang,a28wong\}@uwaterloo.ca}
}

\maketitle

\begin{abstract}
Computer vision based technology is becoming ubiquitous in society. One application area that has seen an increase in computer vision is assistive technologies, specifically for those with visual impairment. Research has shown the ability of computer vision models to achieve tasks such provide scene captions, detect objects and recognize faces. Although assisting individuals with visual impairment with these tasks increases their independence and autonomy, concerns over bias, privacy and potential usefulness arise. This paper addresses the positive and negative implications computer vision based assistive technologies have on individuals with visual impairment, as well as considerations for computer vision researchers and developers in order to mitigate the amount of negative implications.

\end{abstract}

\vspace{-0.2in}
\section{Introduction}
\vspace{-0.1in}
\input{content/intro.tex}

\vspace{-0.1in}
\section{Positive Implications} \label{positive}
\vspace{-0.1in}
\begin{figure}[t]
\begin{center}
   \includegraphics[width=0.45\textwidth]{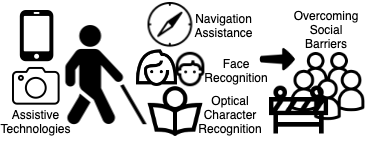}
\end{center}
   \caption{Positive implications: computer vision-based devices allow blind individuals to navigate independently, recognize faces and read text, which helps them overcome social barriers.}
\label{fig:pos}
\end{figure}

\input{content/positive.tex}

\section{Negative Implications} \label{neg}
\vspace{-0.1in}
\begin{table}[]
\begin{center}
\label{my-label}
\begin{tabular}{|
>{\columncolor[HTML]{EFEFEF}}l |l|}
\hline
\multicolumn{2}{|c|}{\cellcolor[HTML]{C0C0C0}\textbf{Negative Implications}} \\ \hline
\cellcolor[HTML]{EFEFEF} & Gender \\ \cline{2-2}
\cellcolor[HTML]{EFEFEF} & Age \\ \cline{2-2}
\multirow{-3}{*}{\cellcolor[HTML]{EFEFEF}Bias} & Race/Ethnicity \\ \hline
\cellcolor[HTML]{EFEFEF} & \begin{tabular}[c]{@{}l@{}}Exploitation of personal \\ information\end{tabular} \\ \cline{2-2}
\cellcolor[HTML]{EFEFEF} & Obtrusiveness of cameras \\ \cline{2-2}
\multirow{-3}{*}{\cellcolor[HTML]{EFEFEF}Privacy} & \begin{tabular}[c]{@{}l@{}}Tradeoff between autonomy \\ and privacy costs\end{tabular} \\ \hline
\cellcolor[HTML]{EFEFEF} & Poor device evaluation \\ \cline{2-2}
\cellcolor[HTML]{EFEFEF} & Age and condition dependent \\ \cline{2-2}
\multirow{-3}{*}{\cellcolor[HTML]{EFEFEF}\begin{tabular}[c]{@{}l@{}}Exclusion in \\ development \\ process\end{tabular}} & \begin{tabular}[c]{@{}l@{}}Inefficiency in development \\ process\end{tabular} \\ \hline
\end{tabular}
\end{center}
\caption{Negative implications: bias in computer vision algorithms, privacy concerns related to data collection and cameras, and exclusion in development process.}
\label{table:neg}
\end{table}

Although the advancement of technology is evident, only a limited number of assistive technology solutions have emerged to make a social or economic impact and improve quality of life. Fundamental challenges, such as those shown in Table \ref{table:neg}, are still be to thoroughly addressed before deploying into assistive technologies.

\label{bias}
\input{content/bias.tex}

\label{privacy}
\input{content/privacy.tex}

\label{exclusion}
\input{content/exclusion.tex}

\section{Design Considerations for Researchers}
\vspace{-0.1in}

\begin{figure}[t]
\begin{center}
   \includegraphics[width=0.45\textwidth]{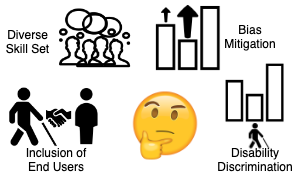}
\end{center}
   \caption{Design considerations for computer vision researchers.}
\label{fig:considerations}
\end{figure}

\input{content/considerations.tex}
\vspace{-0.1in}
{\small
\bibliographystyle{ieee}
\bibliography{fate}
}

\end{document}

%% file: content/intro.tex
In recent years, the rise of deep learning has made previously unsolvable tasks possible. One particular area where deep learning has made tremendous progress is computer vision, such as in image recognition, object detection and image understanding. As computer vision results are becoming more promising, larger issues regarding the use of this technology need to be considered. An important area to consider is assistive technologies for those with visual impairments, as computer vision technologies have the potential to aid in tasks where previous solutions have struggled. 

Although there are positive aspects of computer vision applications, there are also negative aspects that should be addressed. The use of black box artificial intelligence solutions raises many concerns such as fairness and bias of the models~\cite{lloyd-bias}. There are also ethical concerns related to privacy protection as many computer vision models rely on camera input. In addition, the exclusion of certain groups during the development process may also lead to negative aspects of the technology~\cite{riedl-humancenterAI} and as a result, lead to low adoption rates. 

As AI is becoming more ubiquitous, it is crucial to address issues related to the implications of AI, specifically computer vision driven assistive technology towards individuals with visual impairment. The goal of the paper is to review what implications computer vision has on assistive technologies for individuals with visual impairment and considerations for computer vision researchers. The paper will be guided by the following questions:
\begin{itemize}
    \item What are the positive and negative aspects of using computer vision in assistive technologies with respect to the impact on the lives of individuals with visual impairment?
    \item What should researchers consider while conducting computer vision research to reduce negative implications of AI-powered assistive technology on the lives of individuals with visual impairment?
\end{itemize}

%% file: content/positive.tex
Vision impairment and blindness cause a considerable amount of economic and emotional burden for not only the affected persons but also their caregivers and society at large~\cite{kober-burden}. The recent rise in computer vision based assistive technologies show the potential to reduce some burden placed on the individuals, as well as on caregivers and society. By assisting visually impaired individuals with tasks they would otherwise need help in, as shown in Figure \ref{fig:pos}, their level of independence and autonomy are increased.

\textbf{Overcoming social barriers:} One area assistive technologies have become an integral part in the lives of those with visual impairment is overcoming barriers faced in everyday life. These individuals face adversity in all stages of life. For instance, severely visually impaired young people use their assistive technology as more than just a device to overcome environmental barriers but also a means of communication for peers in their school~\cite{exclusion-young}.

\textbf{Face recognition and optical character recognition:} The ever growing presence of smartphones and advancements in computer vision are transforming the accessibility of assistive technologies, allowing individuals to overcome social barriers and have autonomy over when and how they access information. Smartphone applications, such as SeeingAI and Lookout, use auditory cues to assist users in identifying scenes, recognizing faces, reading short text, documents and currency~\cite{seeingAI, lookout}.

\textbf{Navigation assistance:} Individuals with visual impairment also face difficulty localizing themselves in unknown indoor and outdoor environments. Research projects are using cameras and sensors to give directions so these individuals can navigate outdoor and indoor environments independently. For instance, a prototype was developed for guiding the visually impaired across streets in a straight line using a wearable computer-based orientation and wayfinding aid~\cite{2001_ross}. For indoor navigation, Tian et al. developed a proof of concept computer-vision based indoor wayfinding aid that detects doors and elevators, as well as text on signs, to find different rooms~\cite{2013_tian}. 

%% file: content/bias.tex
\textbf{Bias:} In machine learning, bias refers to statistics that lead to a skew and as a result, brings an unjust outcome for a population~\cite{lloyd-bias}. Bias often stems from training data sample sets that are non-representative of the general population. When algorithms are trained with biased data, they are inherently bound to produce skewed results~\cite{buolamwini-bias}.

One of the biggest implications in applying AI systems with bias is the potential for adversely impacting already marginalized groups. In 2012, Klare et al. conducted a study on the influence of gender, race/ethnicity and age on the performance of six different face recognition algorithms, three of which are commercial~\cite{klare-bias}. The results found that there are lower matching accuracies for females than males, Blacks compared to other race/ethnicities, and 18 to 30 year olds compared to other age groups.

In recent years, the low errors rates achieved by facial recognition models led to even more commercialization. However, studies have shown consistent bias in areas of gender, race and age from these commercial models. Buolamwini and Gebru evaluated bias present in three commercial automated facial analysis algorithms from IBM, Microsoft and Megvii with respect to phenotypic subgroups~\cite{buolamwini-bias}. The results showed that there is a significant drop in performance of state of the art models when applied to images of a particular gender and/or ethnicity group. For instance, male subjects were more accurately classified than females and lighter subjects were more accurately classified than darker subjects. All three commercial classifier performed the worst on darker female subjects.

Raji and Buolamwini conducted a second audit of commercial facial analysis models~\cite{raji-bias}. In this study, performances from target companies, ones that were in the first audit, and non-target companies, Amazon and Kairos, are presented. The results showed all targets had the greatest reduction in error rates for female and darker faces. In terms of non-target companies, the performance results were similar to the first audit, with the largest disparity gap between black females and white males.

Although the awareness of disparity improved the facial recognition models from target companies and produced a lower error rate than non-target companies, the commercialization of these models before evaluating biases and potential impacts on protected groups raises a concern.  

%% file: content/privacy.tex
\textbf{Privacy:} As shown in the Section \ref{positive}, computer vision based assistive technologies for the visually impaired allow these individuals to gain independence and autonomy over different aspects of their life. However, these devices also pose privacy risks because of the vast amounts of personal data stored. Although individuals with visual impairment felt that smartphones help them communicate and achieve greater independence, these devices create privacy risks because of the amount of personal data stored. As well, their poor visual acuity makes it hard to safeguard their information, such as if someone is around and eavesdropping~\cite{crandall-privacy}.

Home-monitoring for older adults, who represent majority of those with visual impairment~\cite{who_2018}, reliefs caregivers burden and allows individuals with severe visual impairment to live independently, but the devices for monitoring also store personal data. Studies have found that older adults are willing to have activity monitoring shared with family members and doctors if the collected data is useful, but expressed that the greatest concern is exploitation and misuse of their personal health information~\cite{kirch-privacy}.

Based on the studies, the greatest fear associated with the collection of personal data is the concern that their collected data could end up in the wrong hands and be misused. In addition to the fear of personal information being exploited, the use of cameras is obtrusive and found to elicit greater fears than wearable solutions. In a comparison of four ambient intelligent systems, the camera-based behaviour and emergency detection system was perceived with the greatest fear and highest level of concern~\cite{kirch-privacy}. However, studies have also shown that there is a tradeoff between gained autonomy and privacy costs. Older adults with lower levels of functioning are willing to accept video cameras and tradeoff the privacy lost if camera-based solution could prevent transfer to a long term care facility~\cite{townsend-privacy}.

The different perceptions of privacy over the use of data, as well as the potential benefits of using cameras for home monitoring, suggest that privacy is a complex topic. Understanding the variables that influence privacy concerns and how these concerns can be mediated by potential benefits are important when developing computer vision based assistive technologies.  

%% file: content/exclusion.tex
\textbf{Exclusion in development process:} The main goal of assistive technologies is to improve the lives of end users. However, when the design of form or function of the technology is poor, or when inequality exists between technological accessibility, the lives of those affected can be negatively impacted, as well as perceptions of their abilities~\cite{leo-exclusion}. For instance, a device that has good design, usability and accessibility can be poorly evaluated. The user's lifestyle and aspirations have to be taken into consideration to receive a positive user evaluation~\cite{leo-exclusion}.

The lifestyle and desired function of assistive technologies depend on age and level of adaption to their condition. A predominant want for young disabled people is the significance of being ordinary~\cite{exclusion-young}. No matter the degree of visual impairment, all the participants expressed that inclusion by peers and being ordinary is a big part in their daily lives~\cite{exclusion-young}. In addition to age, how the user has adapted to their condition also impacts the desired functionality of the assistive technology. As users become more accustomed to their condition, they may prefer to perform some activities independently~\cite{2001_ross}. 

Not only does including users during the development process point out which areas to focus on, but also saves development time. When testing the usability of the indoor wayfinding device on blind participants, the researchers found that the participants were able to find doors without any problem since the participants use canes, and realized that text localization and recognition were more useful for indoor navigation~\cite{2013_tian}. By including the users earlier in the development process could have identified that locating doors are not a problem and use the saved time to address text localization and recognition.

%% file: content/considerations.tex
Computer vision has the potential to impact people's lives. However, just algorithmic advances to the accuracies of computer vision models are insufficient for assistive technologies, which interact with and around humans. Recently, the term \textit{human-centered artificial intelligence} is used to refer to intelligent systems that are aware of the interaction with humans and are designed with social responsibility in mind~\cite{riedl-humancenterAI}. As researchers, it is important to uphold society's moral and legal obligations to treat citizens fairly, especially those in protected groups that face discrimination. Figure \ref{fig:considerations} illustrates some considerations to reduce the negative implications mentioned in Section \ref{neg}.

\textbf{Bias mitigation:} One method to uphold fairness is by mitigating bias. For instance, researchers can use tools, such as Google's What-If tool~\cite{what-if} and IBM AI Fairness 360 kit~\cite{aif360}, to analyze and identify unwanted bias in datasets and ML models in order to mitigate such bias. For age, gender and ethnicity, there are different methods to reduce the negative impacts of bias. Das et al. proposed a Multi-Task Convolution Neural Network that employs joint dynamic loss weight adjustments to minimize bias when classifying gender, age and race~\cite{das-bias}. There are also methods to reduce bias at the dataset level. Salimi et al. introduced a database repair algorithm, which uses causal pre-processing to reduce or eliminate sources of discrimination for fair ML~\cite{salimi-bias}.

\textbf{Disability discrimination:} Like age, gender and race, disability status is also a protected characteristic. However, disability discrimination has not been explored in literature~\cite{trewin-bias}. Similar to under-representation of age, gender and racial groups in datasets, as shown in Section \ref{bias}, there is also potential for under-representation of individuals with disabilities. Ways to mitigate disability bias have also not been explored. Compared to gender, race and age, gathering a balanced training dataset is not enough to address the biased outcomes for those with a disability~\cite{trewin-bias}. The many different forms and degrees of disability makes it difficult for a machine learning model to find patterns, form groups and generalize. With the rise of machine learning based assistive technologies, understanding and assessing the impact towards people with disabilities is crucial, especially since disability bias has not been widely explored.

\textbf{Inclusion of end users:} Taking into account where end users will use the assistive technologies, as well as the needs and goals, a task specific training set and appropriate model architecture can allow computer vision based devices to be perceived as useful, allowing individuals to gain independence and autonomy. Based on the studies mentioned in Section \ref{privacy}, users are willing to tradeoff privacy for more autonomy. Thus, by including users in the development process, the devices will be perceived as more useful, gaining more adoption since users are willing to tradeoff privacy concerns to have more independence and autonomy.

\textbf{Diverse skill set: }The ethical implications presented in this paper are difficult to address by just computer vision researchers. Instead, a team with a diverse set of skills is required to address both the positive and negative implications of an assistive technology. The underlying bias in the models can cause protected groups to feel more isolated. Researchers should be aware of the possible biases the dataset and algorithm may have before the system becomes commercialized and interacts with people in everyday context. In addition to bias, the use of cameras raise privacy concerns over who has access to the data stored and the amount of security measures taken to protect personal data. Before deployment, developers should ensure that measures are in place to reduce the chances of data exploitation. By understanding the needs and goals of individuals with visual impairment, designers can effectively address these requirements in the design of the computer vision system, resulting in a more useful device for the end users.